\documentclass[letterpaper]{article} 
\usepackage{aaai2026}  
\usepackage{times}  
\usepackage{helvet}  
\usepackage{courier}  
\usepackage[hyphens]{url}  
\usepackage{graphicx} 
\usepackage{amssymb} 
\usepackage{marvosym} 
\usepackage{amsmath} 
\usepackage{natbib}  
\usepackage{caption} 
\usepackage{algorithm}
\usepackage{algorithmic}

\usepackage{newfloat}
\usepackage{listings}
\DeclareCaptionStyle{ruled}{labelfont=normalfont,labelsep=colon,strut=off} 
\floatstyle{ruled}
\newfloat{listing}{tb}{lst}{}
\floatname{listing}{Listing}

\title{Perception Before Reasoning: Two-Stage Reinforcement Learning for Visual Reasoning in Vision-Language Models}
\author{
    Yan Chen\textsuperscript{\rm 1},
    Long Li\textsuperscript{\rm 2},
    Teng Xi\textsuperscript{\rm 2},
    Long Zeng\textsuperscript{\rm 1}\textsuperscript{\Letter},
    Jingdong Wang\textsuperscript{\rm 2}
}
\affiliations{
    \textsuperscript{\rm 1}Tsinghua University \textsuperscript{\rm 2}Baidu Inc.\\

}

\begin{document}

\maketitle
\begin{abstract}

Reinforcement learning (RL) has proven highly effective in eliciting the reasoning capabilities of large language models (LLMs). Inspired by this success, recent studies have explored applying similar techniques to vision-language models (VLMs), aiming to enhance their reasoning performance. However, directly transplanting RL methods from LLMs to VLMs is suboptimal, as the tasks faced by VLMs are inherently more complex. Specifically, VLMs must first accurately perceive and understand visual inputs before reasoning can be effectively performed. To address this challenge, we propose a two-stage reinforcement learning framework designed to jointly enhance both the perceptual and reasoning capabilities of VLMs. To mitigate the vanishing advantage issue commonly observed in RL training, we first perform dataset-level sampling to selectively strengthen specific capabilities using distinct data sources. During training, the first stage focuses on improving the model's visual perception through coarse- and fine-grained visual understanding, while the second stage targets the enhancement of reasoning abilities. After the proposed two-stage reinforcement learning process, we obtain PeBR-R1, a vision-language model with significantly enhanced perceptual and reasoning capabilities. Experimental results on seven benchmark datasets demonstrate the effectiveness of our approach and validate the superior performance of PeBR-R1 across diverse visual reasoning tasks. 
\end{abstract}

\begin{links}
    \link{Project Page}{https://github.com/cythu/PeBR-R1}
\end{links}

\renewcommand{\thefootnote}{} 
\footnotetext{
\Letter~Corresponding author. \\
\hspace*{0em}
}
\renewcommand{\thefootnote}{\arabic{footnote}} 

\section{Introduction}

Reinforcement learning (RL) has become a key technique for aligning large language models (LLMs) with human preferences~\cite{ouyang2022training, gpt4, rlaif, christiano2017deep}. Recent methods such as GRPO~\cite{deepseekmath,deepseekr1} and DAPO~\cite{dapo} further improve reasoning with more stable optimization. Building on this, RL has been applied to vision-language models (VLMs) to enhance multimodal reasoning~\cite{openvlthinker,perceptionr1,vlrethinker,vlmr1,r1sharevl,visionr1,thinklitevl}. However, most approaches directly adopt RL paradigms from LLMs without adapting to visual inputs, leading to visual neglect\cite{qwenlookagain,vicrit}, where models overfit language priors while ignoring image content. Since mainstream VLMs are trained under weak language supervision (e.g., captioning, dialogue), they struggle with fine-grained perception tasks such as object relations and spatial reasoning\cite{shapeblind,slowperception}. These limitations highlight the need for RL strategies that explicitly enhance both perception and reasoning.

\begin{figure}[t]
\centering
\includegraphics[width=\columnwidth]{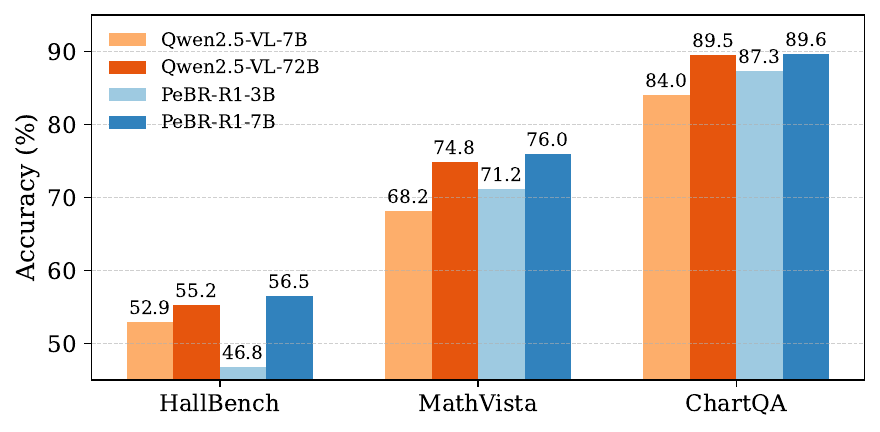}
\caption{Performance comparison of PeBR-R1 with existing open-source VLMs.}
\vspace{-1.0em} 
\label{fig1}
\end{figure}

To enhance the performance of vision-language models (VLMs) on multimodal reasoning tasks, we propose a two-stage reinforcement learning framework that jointly optimizes the model’s perceptual and reasoning abilities. By carefully designing reward signals, our approach effectively guides the model to attend to key visual regions and semantic concepts closely related to the given question, thereby progressively improving its multimodal reasoning capability. As illustrated in Figure~\ref{fig1}, our 7B model even outperforms certain open-source models with up to 72B parameters on multiple vision-language reasoning benchmarks, demonstrating the effectiveness of the proposed framework.

To achieve such performance gains, we design a refined training procedure to ensure both the stability and efficacy of the reinforcement learning process. First, to provide the model with initial perceptual and reasoning capabilities, we perform a warm-up stage through supervised fine-tuning (SFT) on the Mulberry-260K dataset \cite{mulberry}. However, we observe a noticeable performance drop on certain reasoning tasks after SFT. To mitigate this issue, we augment the training set with additional targeted examples that specifically address the model’s weaknesses. Second, prior to the main reinforcement learning phase, we introduce a sample filtering mechanism to alleviate performance degradation caused by vanishing advantage signals, which may arise from overly consistent reward patterns. Specifically, for each question, we perform 8 independent rollouts to generate diverse model responses. Based on the number of correct answers, each question is categorized into one of three subsets: Easy cases (all 8 answers are correct), Medium cases (partially correct), and Hard cases (all incorrect).

In the first reinforcement learning stage, we focus explicitly on enhancing the model’s visual perception capabilities, without directly optimizing for the correctness of final answers. To ensure that the model is guided by reliable perceptual signals, we exclusively select Easy cases for training, thereby minimizing the risk of reinforcing spurious correlations between perception and incorrect outputs. We extract the \texttt{Image Description} section from model outputs to construct two complementary reward signals. The first is based on FGCLIP~\cite{fgclip}, which measures coarse-grained alignment between the generated descriptions and the input image. The second leverages a set of fine-grained semantic keywords generated by a pretrained vision-language model (Seed1.5-VL)~\cite{seed1.5vl} and manually curated to ensure quality and relevance, thereby encouraging the model to recognize key visual concepts essential for solving the task.

The second stage of reinforcement learning is dedicated to enhancing the model’s reasoning abilities. To mitigate the issue of vanishing advantages and provide stable gradient updates, we use Medium cases as training samples in this stage. We employ rule-based reward signals to improve the logical consistency and reliability of the model’s outputs. These reward signals include format correctness rewards, which encourage structured responses, and accuracy rewards that ensure the correctness of final answers. For optimization in the two-stage reinforcement learning framework, we adopt the Group-based Relative Policy Optimization (GRPO) method, leveraging normalized advantage estimation across multiple candidate responses to stabilize policy updates. The previously described sample filtering and stage-control strategies effectively prevent interference from ineffective learning signals.

Our main contributions are as follows:
\begin{itemize}
    \item We propose a two-stage reinforcement learning framework that progressively enhances visual perception and multimodal reasoning abilities.
    \item We propose a GRPO-based visual Perception Reinforcement Learning approach, which integrates image-text consistency and fine-grained keyword alignment rewards, and effectively enhances the model’s capabilities in object recognition, numerical understanding, attribute comprehension, and spatial relation modeling.
    \item Through a two-stage reinforcement learning process, we obtain PeBR-R1, a vision-language model with significantly enhanced perceptual and reasoning capabilities. Compared to existing methods, PeBR-R1 demonstrates consistently superior performance across a range of visual perception benchmarks.
\end{itemize}

\section{Related Work}
\subsubsection{Reinforcement Learning for Large Language Models}
Recent advances highlight the critical role of reinforcement learning (RL) in aligning large language models (LLMs) with human preferences. A typical RL pipeline involves training a reward model, scoring outputs, and optimizing the policy using algorithms such as PPO~\cite{ppo}. RLHF~\cite{christiano2017deep}, popularized by InstructGPT~\cite{ouyang2022training} and adopted in GPT-4~\cite{gpt4}, remains the standard, but suffers from reward bias, high annotation costs, and instability. To reduce reliance on human feedback, RLAIF~\cite{rlaif} replaces it with LLM-based preference judgments~\cite{bai2022constitutional,bai2022training}, as demonstrated in Claude 3.5 Sonnet~\cite{claude3_5_sonnet_2024} and Starling-7B~\cite{starling}. Recent methods such as DPO~\cite{dpo} improve efficiency by directly optimizing pairwise preferences with contrastive loss. Building on this, GRPO~\cite{deepseekmath,deepseekr1} and DAPO~\cite{dapo} offer further gains—GRPO avoids value function training via group-wise comparisons and KL regularization, while DAPO introduces dynamic sampling and stable updates to accelerate convergence. These methods underscore the importance of RL in enhancing both performance and alignment of LLMs.

\subsubsection{Reinforcement Learning for Vision-Language Models} 
Inspired by the success of reinforcement learning (RL) in large language models (LLMs), recent efforts have extended RL to multimodal settings to enhance vision-language model (VLM) reasoning. Vision-R1~\cite{visionr1} introduces a cold-start pipeline with a 200K multimodal CoT dataset, using GRPO under strict format constraints. R1-VL~\cite{r1vl} proposes StepGRPO with step-wise rewards to improve logical consistency. R1-ShareVL~\cite{r1sharevl} alleviates sparse rewards by expanding question space and sharing reasoning signals. VL-Rethinker~\cite{vlrethinker} promotes slow thinking via selective replay and rethinking. OpenVLThinker~\cite{openvlthinker} interleaves supervised fine-tuning (SFT) and RL to refine reasoning. VLM-R1~\cite{vlmr1} applies rule-based RL to vision tasks, focusing on stability and reward hacking. ThinkLite-VL~\cite{thinklitevl} leverages MCTS to mine hard cases, achieving strong results with limited data. Visionary-R1~\cite{visionaryr1} encourages visual grounding via a caption–reason–answer format and LLM-based caption rewards, mitigating shortcut behaviors and boosting generalization. Despite these advances, most methods still emphasize textual reasoning, underutilizing visual information and highlighting the need for better grounded RL frameworks.

\subsubsection{Visual Perception in Vision-Language Models}
To enhance the visual understanding capabilities of vision-language models (VLMs), recent studies address this gap with targeted solutions. \cite{bringreasontovision} proposes model merging as a training-free way to inject reasoning from LLMs into VLMs by fusing cross-modal parameters, showing that perception and reasoning are localized in early and later layers, respectively. \cite{shapeblind} finds that VLMs lack structured geometric reasoning, often failing to identify basic shapes. Cambrian-1~\cite{cambrian1} adopts a vision-centric design, benchmarking 20+ vision encoders and introducing CV-Bench and the Spatial Vision Aggregator (SVA) to enhance visual grounding. \cite{slowperception} mimics human step-by-step perception to handle geometric figures via decomposition and reconstruction. Perception-R1~\cite{perceptionr1} applies Group Relative Policy Optimization (GRPO) to learn better perception policies, achieving strong benchmark results. DINO-R1~\cite{dinor1} introduces Group Relative Query Optimization (GRQO), a reinforcement-based method tailored for models like Grounding-DINO\cite{groundingdino}, enabling query-level supervision and improved contextual reasoning. Together, these works show that VLMs’ visual modules remain under-optimized, and approaches like model merging, architectural innovation, and reinforcement learning are key to advancing visual perception and reasoning.

\begin{figure*}[t]
\centering
\includegraphics[width=\textwidth]{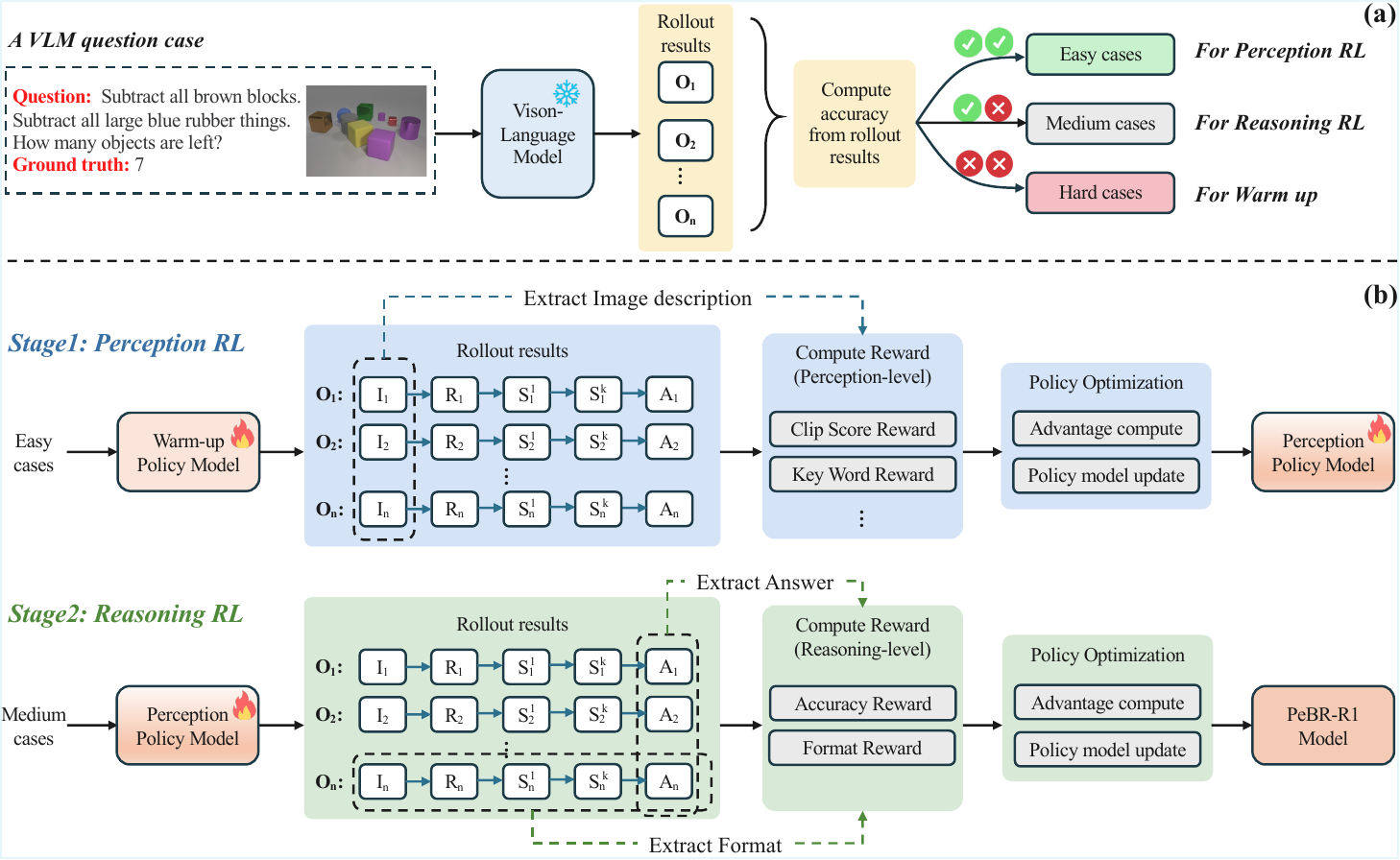} 
\caption{
(a) Overview of question sampling and categorization.
Each question is passed to the vision-language model (VLM) for n independent rollouts. Based on the number of correct responses, questions are categorized into three types: Easy cases (all correct), Medium cases (partially correct), and Hard cases (all incorrect).
(b) Overview of the two-stage reinforcement learning framework.
The model output is parsed into four components: Image Description (I), Rationale (R), Step-by-step Thinking ($S^1$ to $S^k$), and Final Answer (A).
In Stage 1: Perception RL, we use Easy cases to train the warm-up policy model by extracting image descriptions, computing rewards and advantages, and updating the policy to obtain the perception policy model.
In Stage 2: Reasoning RL, we continue training the perception policy model on Medium cases using the final answer and output format to compute rewards and advantages, and optimize the policy to obtain the final PeBR-R1 model.
}
\vspace{-0.7em} 
\label{fig2}
\end{figure*}

\section{Preliminaries}
\subsubsection{Group Relative Policy Optimization (GRPO)}

Group Relative Policy Optimization (GRPO) is a reinforcement learning algorithm based on Proximal Policy Optimization (PPO), specifically designed to enhance reasoning capabilities in large language models. The core idea is to generate a group of candidate responses $\{o_i\}_{i=1}^G$ for each question $q$, and compute the policy ratio as $r_i(\theta) = \pi_\theta / \pi_{\theta_{\text{old}}}$. Based on the reward $R_i$ for each response, GRPO computes a normalized advantage function $A_i$ using the group-level mean and standard deviation: 
\begin{equation}
A_i = \frac{R_i - \mathrm{mean}(\{R_1, R_2, \ldots, R_G\})}{\mathrm{std}(\{R_1, R_2, \ldots, R_G\})}
\label{advantage compute}
\end{equation}

To ensure training stability, GRPO applies a clipping strategy that restricts $r_i(\theta)$ to the interval $(1 - \varepsilon, 1 + \varepsilon)$, preventing overly large policy updates. The core optimization term becomes $\min \left( r_i(\theta) A_i, \, \text{clip}(r_i(\theta), 1 - \varepsilon, 1 + \varepsilon) A_i \right)$. Additionally, GRPO introduces a KL regularization term $D_{\mathrm{KL}}(\pi_\theta \| \pi_{\mathrm{ref}})$ to constrain the divergence between the current policy and a reference policy (typically the initial policy), mitigating the risk of drifting too far from the model's original capabilities. The final optimization objective of GRPO is to maximize:

\begin{align}
&J_{\mathrm{GRPO}}(\theta)
= \mathbb{E}_{q,\, \{o_i\}} \bigg[
\frac{1}{G} \sum_{i=1}^G \min\big(
r_\theta(o_i \mid q) A_i, \notag \\
&\quad\mathrm{clip}(r_\theta(o_i \mid q), 1{-}\varepsilon, 1{+}\varepsilon) A_i
\big)
- \beta D_{\mathrm{KL}}(\pi_\theta \,\|\, \pi_{\mathrm{ref}}) \bigg]
\end{align}

\section{Methodology}

In this section, we first describe the preparation steps for reinforcement learning, including dataset sampling and model warm-up. We then present our two-stage reinforcement learning framework, which comprises a perception-level stage and a reasoning-level stage, each designed to enhance distinct aspects of the model’s capabilities.

\subsection{Dataset Sample}

\subsubsection{Model Degradation}

During the GRPO reinforcement learning process, the advantage signal may vanish, primarily due to reward uniformity within a group. To investigate this issue, we analyze the optimization gradient. For simplicity, the policy ratio is assumed to satisfy ${\pi_{\theta}} / {\pi_{\theta_{\text{old}}}} \in (1 - \varepsilon,\; 1 + \varepsilon)$. Since the objective is to maximize $J_{\mathrm{GRPO}}(\theta)$, this is equivalent to minimizing the loss $\mathcal{L}_{\pi_\theta} = -J_{\mathrm{GRPO}}(\theta)$. The corresponding gradient is given as follows:

\begin{align}
\nabla_\theta \mathcal{L}_{\pi_\theta}
&= -\mathbb{E}_{q,\, \{o_i\}} \bigg[
\frac{1}{G} \sum_{i=1}^G 
r_\theta(o_i \mid q)\, \nabla_\theta \log \pi_\theta(o_i \mid q) \cdot A_i \notag \\
&\quad - \beta \nabla_\theta D_{\mathrm{KL}}\left(\pi_\theta \,\|\, \pi_{\mathrm{ref}}\right) \bigg]
\end{align}

When all rewards $R$ within a group are identical (e.g., all 0 or all 1), the advantage $A_i$ computed by Eq.~\eqref{advantage compute} becomes zero. In this case, the optimization gradient simplifies to:

\begin{alignat}{2}
&\nabla_\theta \mathcal{L}_{\pi_\theta}
= \beta \nabla_{\theta} D_{\mathrm{KL}}(\pi_\theta \| \pi_{\mathrm{ref}})
\end{alignat}

At this point, the reinforcement learning signal vanishes, leaving only the KL penalty gradient. This may bias the optimization trajectory toward the reference model, potentially leading to model degradation.
\subsubsection{Dataset Sampling}
To prevent model degradation and ensure stable gradient descent during reinforcement learning, we sample the dataset before each reinforcement learning stage. As shown in Figure 2, sampling is guided by the number of correct responses: the model is prompted to generate 8 responses per question, and based on the number of correct responses, we divide the dataset into three subsets:
\begin{itemize}
    \item \textbf{Easy cases}: All 8 generations are correct.
    \item \textbf{Medium cases}: Partially correct (i.e., 1 to 7 correct).
    \item \textbf{Hard cases}: All 8 generations are incorrect.
\end{itemize}

In the Perception Reinforcement Learning (Perception RL) stage, which aims to enhance the model’s perception ability, we utilize the Easy cases subset. This design prevents the model from encountering samples where visual perception is correct but the final answer is incorrect, which could impede early-stage learning. In the Reasoning Reinforcement Learning (Reasoning RL) stage, we leverage the Medium cases subset to provide stable learning signals and promote the model’s reasoning ability through gradient-based optimization. In the warm-up stage, we utilize a portion of the Hard cases as part of the training dataset. The detailed methodology will be introduced in the next section.

\begin{figure*}[t]
\centering
\includegraphics[width=\textwidth]{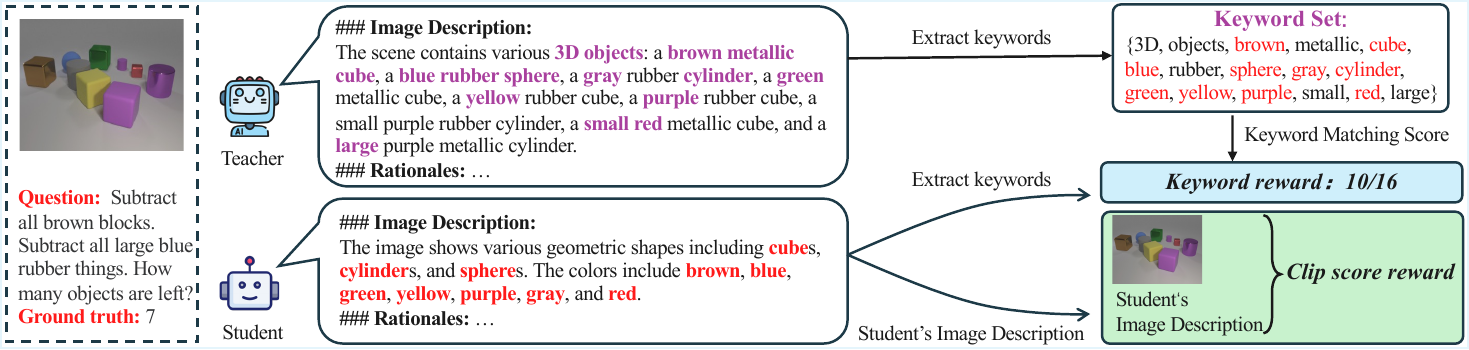} 

\caption{Perception RL Reward Computation. The keyword reward is computed by extracting keywords from teacher-generated image description to construct a reference keyword set, followed by calculating the proportion of matched keywords in the student-generated image description. The CLIP reward is obtained by evaluating the CLIP similarity between the student-generated image description and the corresponding input image, with the resulting score subsequently scaled.}
\vspace{-0.7em} 
\label{fig3}
\end{figure*}

\subsection{Warm-up via Supervised Fine-tuning}
To equip the base model with initial visual understanding and reasoning capabilities, we first conducted supervised fine-tuning (SFT) using the Mulberry dataset. However, we observed a notable drop in accuracy on certain question types after SFT. To address this, we augmented the training data with six additional categories from public datasets, thereby strengthening the model's supervised learning foundation. Since reinforcement learning (RL) relies on the model’s own outputs for optimization, it struggles to learn effectively from Hard cases where all responses are incorrect. To mitigate this, we curated a subset of such cases and converted them into SFT-style data through automated filtering and manual verification. This strategy aims to enhance the model’s generalization ability prior to RL training. In total, we added approximately 39K high-quality SFT samples to improve robustness on difficult instances and ensure stable training. The model’s output after SFT follows a structured format beginning with:

\vspace{-0.7em} 
\begin{quote}
\begin{verbatim}
### Image Description:
### Rationales:
### Let's think step by step.
### Step 1:
### Step 2:
...
### The final answer is: 
\end{verbatim}
\end{quote}
\vspace{-0.7em}

\subsection{Perception Reinforcement Learning}

After completing data sampling and model warm-up, we enhance the model’s visual perception ability using the Easy cases subset. Specifically, we extract the \texttt{Image Description} section from the model-generated responses as the evaluation target for perception-level rewards. For coarse-grained visual perception, we employ FGCLIP~\cite{fgclip}, a vision-language pretrained model tailored for fine-grained image-text alignment tasks. FGCLIP demonstrates strong capability in capturing semantic correspondence between visual and textual content. To quantify the alignment quality, we compute the CLIP similarity score between the image \( I_m \) and the generated description \( I_D \), and define the visual alignment reward based on a predefined similarity threshold \( \tau_{\text{clip}} \), as follows:

\begin{equation}
\text{clipscore} = \text{sim}(I_m, I_D)
\end{equation}

\begin{equation}
r_{\text{clip}} =
\begin{cases}
1, & \text{if } \text{clipscore} \geq \tau_{\text{clip}} \\
\text{clipscore} / \tau_{\text{clip}}, & \text{otherwise}
\end{cases}
\end{equation}

For fine-grained visual understanding, As illustrated in Figure~\ref{fig3}, we introduce a teacher-guided keywords extraction and reward mechanism to enhance the model’s fine-grained visual perception. Specifically, we leverage the supervised fine-tuned teacher model Seed1.5-VL\cite{seed1.5vl} to generate structured responses for each sample in the Easy cases subset. From these responses, we extract semantically critical components from the \texttt{Image Description} section—focusing on key perceptual elements such as object recognition, numerical understanding, attribute comprehension, and spatial relation modeling. To ensure the quality of the extracted keywords, we apply a series of filtering steps: removing redundant items, discarding samples with empty keyword sets, and excluding samples with incorrect final answers. This results in a curated reference keyword set $K = \{k_1, k_2, \ldots, k_n\}$.
To evaluate the alignment between the policy model's generated image descriptions and the reference set, we define a keyword reward. Let $\hat{K}$ denote the set of keywords extracted from the policy-model-generated \texttt{Image Description}. The reward is computed as the ratio of matched keywords:
\begin{equation}
r_{\text{keyword}} = \frac{|\hat{K} \cap K|}{|K|}
\end{equation}
Finally, to ensure that the model outputs follow the expected format and to prevent excessively long responses, we include a format reward and a length penalty. The length penalty is computed as $\min(1.0, L_{\text{expected}} / (L_{\text{actual}} + \epsilon))$, where $L_{\text{expected}}$ and $L_{\text{actual}}$ denote the expected and actual response lengths, respectively. The overall reward used in this stage is defined as:

\begin{equation}
r_1 = \alpha_1 \cdot r_{\text{clip}} + \beta_1 \cdot r_{\text{keyword}} + \gamma_1 \cdot r_{\text{format}} + \delta_1 \cdot r_{\text{length}}
\end{equation}

\subsection{Reasoning Reinforcement Learning}
After completing perception-stage reinforcement learning, the model’s visual capabilities have been substantially improved. We then shift focus to enhancing its reasoning and problem-solving abilities. In this stage, we employ a rule-based reward strategy based on the GRPO algorithm to guide the model toward generating well-structured and accurate solutions. To ensure stable gradient updates, we utilize the Medium cases set for training. The reward function combines a format reward, which encourages the model to follow the chain-of-thought structure, and an accuracy reward, which evaluates the correctness of the model’s final answer. The overall reward used in this stage is defined as:
\begin{equation}
r_2 = \alpha_2 \cdot r_{\text{accuracy}} + \beta_2 \cdot r_{\text{format}}
\end{equation}

The complete training process, including both the perception and reasoning stages, is summarized in Algorithm~\ref{alg:grpo_twostage}.

\begin{algorithm}[tb]
\caption{Two-Stage Reinforcement Learning for Vision-Language Models}
\label{alg:grpo_twostage}
\textbf{Input}: Pre-trained VLM policy $\pi_\theta$; SFT dataset $\mathcal{D}_{\text{SFT}} = \{(Q_n, \tau_n)\}_{n=1}^{N_0}$; RL dataset $\mathcal{D}_{\text{RL}} = \{(Q_n)\}_{n=1}^{N_1}$\\
\textbf{Output}: Optimized policy $\pi_\theta$ with enhanced perception and reasoning abilities

{\raggedright\textbf{Stage 0: Supervised Warm-up}\par\vskip2pt}  
\begin{algorithmic}[1]
\FOR{$\text{iter} = 1$ to $N_0$}
    \STATE Sample $(Q, \tau) \sim \mathcal{D}_{\text{SFT}}$
    \STATE Update $\pi_\theta$ via supervised loss $\mathcal{L}_{\text{SFT}}$
\ENDFOR
\end{algorithmic}

\vskip2pt
{\raggedright\textbf{Stage 1–2: GRPO Reinforcement Learning}\par\vskip2pt}
\begin{algorithmic}[1]
\FOR{stage $\in$ \{Perception, Reasoning\}}
    \STATE Select training subset $\mathcal{D}_{\text{stage}} \subset \mathcal{D}_{\text{RL}}$\\
\quad\quad (e.g., Easy or Medium cases)
    \FOR{$\text{iter} = 1$ to $N_{\text{stage}}$}
        \STATE Sample $Q \sim \mathcal{D}_{\text{stage}}$
        \STATE Generate group of responses $\{c_i\}_{i=1}^M \sim \pi_\theta$
        \STATE Compute stage-specific rewards $\{r_i\}_{i=1}^M$
        \STATE Compute group-relative advantages $\{\hat{A}_i\}_{i=1}^M$
        \STATE Update $\pi_\theta$ via GRPO loss
    \ENDFOR
\ENDFOR
\end{algorithmic}
\end{algorithm}

\begin{table*}[t]
\centering
\begin{tabular}{lcccccccc}
\hline
\textbf{Model} & \textbf{\begin{tabular}[c]{@{}c@{}}\textbf{RL} \\ \textbf{Method}\end{tabular}} & \textbf{\begin{tabular}[c]{@{}c@{}}\textbf{MathVista}\\ \textbf{(testmini)}\end{tabular}} & \textbf{\begin{tabular}[c]{@{}c@{}}\textbf{MathVision}\\ \textbf{(full)}\end{tabular}} & \textbf{DynaMath} & \textbf{MMStar} & \textbf{\begin{tabular}[c]{@{}c@{}}AI2D\\(w.M.)\end{tabular}} & \textbf{\begin{tabular}[c]{@{}c@{}}\textbf{HallBench}\\ \textbf{(avg)}\end{tabular}} & \textbf{ChartQA} \\ \hline
\multicolumn{9}{c}{\textbf{Closed Source MLLMs}} \\ \hline
GPT-4o & - & 63.8 & 30.3 & 63.7 & 63.9 & 84.6 & 55.0 & 85.7 \\
Claude-3.5 Sonnet & - & 67.7 & - & 64.8 & 62.2 & 81.2 & 55.0 & 90.8 \\ \hline
\multicolumn{9}{c}{\textbf{Open Source General MLLMs}} \\ \hline
Qwen2.5-VL-3B & RLHF & 62.3 & 21.2 & \underline{42.8}\textsuperscript{*} & 55.9 & \textbf{81.6} & \underline{46.3} & \underline{84.0} \\
Qwen2.5-VL-7B & RLHF & 68.2 & 25.1 & \underline{51.3}\textsuperscript{*} & 63.9 & \underline{83.9} & 52.9 & \underline{87.3} \\
Qwen2.5-VL-72B & RLHF & 74.8 & 38.1 & - & 70.8 & 88.7 & 55.2 & 89.5 \\
InternVL2.5-8B & RLHF & 64.4 & 19.7 & - & 62.8 & 84.5 & 50.1 & 84.8 \\
InternVL2.5-78B & RLHF & 72.3 & 32.2 & - & 69.5 & 89.1 & 57.4 & 88.3 \\ \hline
\multicolumn{9}{c}{\textbf{Open Source Reasoning MLLMs}} \\ \hline
Visionary-R1-3B & P\&R-RL & \underline{69.4} & \underline{24.7} & - & \textbf{66.5} & - & - & - \\
R1-VL-7B & R-RL & 63.5 & 24.7 & 45.2 & 60.0 & - & \underline{54.7} & 83.9 \\
ThinkLite-VL-7B & R-RL & 75.1 & - & - & 65.0 & 83.6 & - & - \\
OpenVLThinker-7B & R-RL & 70.2 & 25.3 & - & - & - & - & - \\
Vision-R1-7B & R-RL & 73.5 & - & - & - & - & - & - \\
VL-Rethinker-7B & R-RL & 74.9 & \underline{32.3} & - & - & - & - & - \\
R1-ShareVL-7B & R-RL & \underline{75.4} & 29.5 & - & \underline{67.0} & \textbf{84.5} & - & - \\ \hline
PeBR-R1-3B (Ours) & P+R-RL & \textbf{71.2} & \textbf{28.1} & \textbf{52.0} & \underline{60.5} & \underline{81.3} & \textbf{46.8} & \textbf{86.4} \\
PeBR-R1-7B (Ours) & P+R-RL & \textbf{76.0} & \textbf{32.7} & \textbf{56.9} & \textbf{67.1} & 83.3 & \textbf{56.5} & \textbf{89.6} \\ \hline
\end{tabular}

\caption{Main Results. Comparison with existing open- and closed-source models. P\&R-RL denotes optimization of perception and reasoning in a single stage; R-RL denotes reasoning-only training; and P+R-RL denotes a two-stage approach with separate optimization of perception and reasoning. For models with 3B or 7B parameter sizes, bold and underlined values indicate the top-1 and top-2 performance, respectively. * indicates results reproduced by us.}
\vspace{-0.7em} 
\label{table1}
\end{table*}

\section{Experiments}

\subsection{Experimental Setup}
\subsubsection{Dataset}

During the supervised fine-tuning (warm-up) stage, we adopt the Mulberry-260K dataset as the primary SFT corpus. To address performance degradation observed in certain capabilities after SFT, we introduce an additional dataset of 39K samples. This supplementary set includes datasets from CLEVR-Math~\cite{clevermath}, DVQA~\cite{dvqa}, TabMWP~\cite{tabmwp}, VQA2.0~\cite{vqa2.0}, Super-CLEVR~\cite{superclevr}, GeoQA+\cite{geoqa+}, and a subset of Hard cases collected via our sampling strategy. In the reinforcement learning phase, we sample from various open-source vision-language datasets\cite{vlrethinker,thinklitevl} and generate candidate responses using the Seed1.5-VL model. We apply keyword-based filtering and manual verification to discard samples with incorrect answers or redundant keywords, resulting in a high-quality reinforcement learning dataset of 110K examples. In the first stage of reinforcement learning (Perception RL), we leverage 36K filtered Easy cases to improve the model’s visual grounding and image description capabilities. In the second stage (Reasoning RL), we use 49K Medium cases to further enhance the model’s logical reasoning and problem-solving performance.

\begin{table}[t]
\centering
\begin{tabular}{l|c}
\hline
RL Method                           & \multicolumn{1}{l}{MathVista} \\ \hline
Warm-up Model (Baseline)            &                      69.8         \\
Reasoning RL                        &                      73.9         \\
Perception\&Reasoning RL            &                      72.7         \\
Perception RL + Reasoning RL (Ours) &                      76.0         \\ \hline
\end{tabular}
\caption{Ablation study of reinforcement learning strategies based on the warm-up Qwen2.5-VL-7B model.}
\vspace{-0.7em} 
\label{table2}
\end{table}

\subsubsection{Training Setup}

We use Qwen2.5-VL-3B and Qwen2.5-VL-7B\cite{qwen25vl} as our base models. In the warm-up stage, we set the batch size to 128, with learning rates of 2e-5 for the 3B model and 1e-5 for the 7B model. In the reinforcement learning stage, for each question, we set both the number of sampling rollouts and training rollouts to 8. The sampling temperature is set to 1.2 to encourage the model to explore diverse reasoning paths.

\subsection{Main Results}
We conduct a comprehensive evaluation of our proposed PeBR-R1 model across seven representative multimodal reasoning benchmarks, including MathVista\cite{mathvista},  MathVision\cite{mathvision}, DynaMath\cite{dynamath}, MMStar\cite{mmstar}, AI2D\cite{ai2d}, HallusionBench\cite{hallusionbench} and ChartQA\cite{chartqa}. The main results are presented in Table~\ref{table1}. We first compare PeBR-R1 with the open-source base model series Qwen2.5-VL. For this family of general-purpose multimodal large language models (MLLMs), our PeBR-R1-3B and PeBR-R1-7B achieve accuracy improvements of +8.9\% and +7.8\% respectively on MathVista, significantly outperforming their counterparts of similar size and even surpassing larger models. For instance, on MathVista benchmark, PeBR-R1-3B outperforms InternVL2.5-8B\cite{internvl2.5} by 6.8\%, while PeBR-R1-7B exceeds Qwen2.5-VL-72B and InternVL2.5-78B by +1.2\% and +3.7\%, respectively. Moreover, we compare PeBR-R1 with several open-source multimodal reasoning models. As shown in Table~\ref{table1}, PeBR-R1 consistently achieves better performance across most benchmarks. Specifically, PeBR-R1-7B outperforms R1-ShareVL-7B by +0.6\%, and PeBR-R1-3B surpasses Visionary-R1-3B by +1.8\% on MathVista. Finally, we compare PeBR-R1 with several powerful closed-source MLLMs. The results show that our model outperforms both GPT-4o\cite{gpt4o} and Claude-3.5 Sonnet on multiple benchmarks. These results highlight the effectiveness of our two-stage reinforcement learning framework in improving multimodal reasoning.

\begin{table}[t]
\centering
\begin{tabular}{c c c|c}
\hline
Warm-up & Perception RL & Reasoning RL & MathVista \\ \hline
        &               &              &    68.2       \\
\checkmark &           &              &      69.8     \\
\checkmark & \checkmark &              &     71.4      \\
\checkmark &           & \checkmark   &      74.3     \\
\checkmark & \checkmark & \checkmark   &      76.0     \\ \hline
\end{tabular}
\caption{Ablation study on multi-stage training strategies based on Qwen2.5-VL-7B.}
\label{table3}
\end{table}

\begin{table}[t]
\centering
\begin{tabular}{l|cc}
\hline
Method & Mean Length & MathVista \\ \hline
w/o Length Penalty & 485.6 & 69.1 \\
w/ Length Penalty  & 359.2 & 71.4 \\ \hline
\end{tabular}
\caption{Impact of length penalty on response length and MathVista accuracy in the Perception RL stage.}
\vspace{-0.7em} 
\label{table4}
\end{table}

\subsection{Ablation Study}
To evaluate the effectiveness of our two-stage reinforcement learning framework in enhancing both perception and reasoning in Vision-Language Models (VLMs), we conduct ablation studies. Unlike Reasoning RL, which focuses solely on reasoning, and single-stage Perception \& Reasoning RL, which jointly trains both abilities, our method separates training into two distinct phases, enabling targeted learning for each capability. We use a supervised warm-up Qwen2.5-VL-7B model as the baseline and evaluate on the MathVista benchmark. All methods share the same dataset (Easy and Medium cases), reward setup, and training parameters to ensure fairness. As shown in Table~\ref{table2}, while Reasoning RL brings some improvement, it underperforms compared to our approach due to insufficient perception learning. Notably, the Perception \& Reasoning RL strategy performs even worse, as simultaneous reinforcement introduces training ambiguity and hinders focused optimization. In contrast, our two-stage method explicitly separates perception and reasoning phases, avoiding interference and enabling precise skill acquisition. The results confirm the superiority of our approach in complex multimodal reasoning tasks.

We divide the training into three stages: warm-up, Perception RL and Reasoning RL. To assess each stage’s contribution, we conduct ablation experiments using the Qwen2.5-VL-7B model. The warm-up stage uses supervised fine-tuning (SFT), while the reinforcement stages use Easy and Medium cases, respectively. MathVista is used for evaluation. As shown in Table~\ref{table3}, the warm-up stage boosts baseline performance to 69.8\% by providing basic perception and reasoning abilities. Both reinforcement stages individually yield further gains, and their combination achieves the best result at 76.0\%. These results show that our method significantly improves performance on complex vision-language reasoning tasks.

During the Perception RL stage, the model tends to produce excessively long outputs in order to maximize perception-related rewards. This behavior can lead to output loops or redundancies, thereby hindering the effectiveness of subsequent reasoning-stage training. To mitigate this issue, we introduce a length penalty during the first stage to regularize output verbosity. As shown in Table~\ref{table4}, the application of the length penalty significantly reduces the average output length and concurrently improves solution accuracy on the MathVista benchmark.

\subsection{Training Visualization}
\begin{figure}[t]
\centering
\includegraphics[width=\columnwidth]{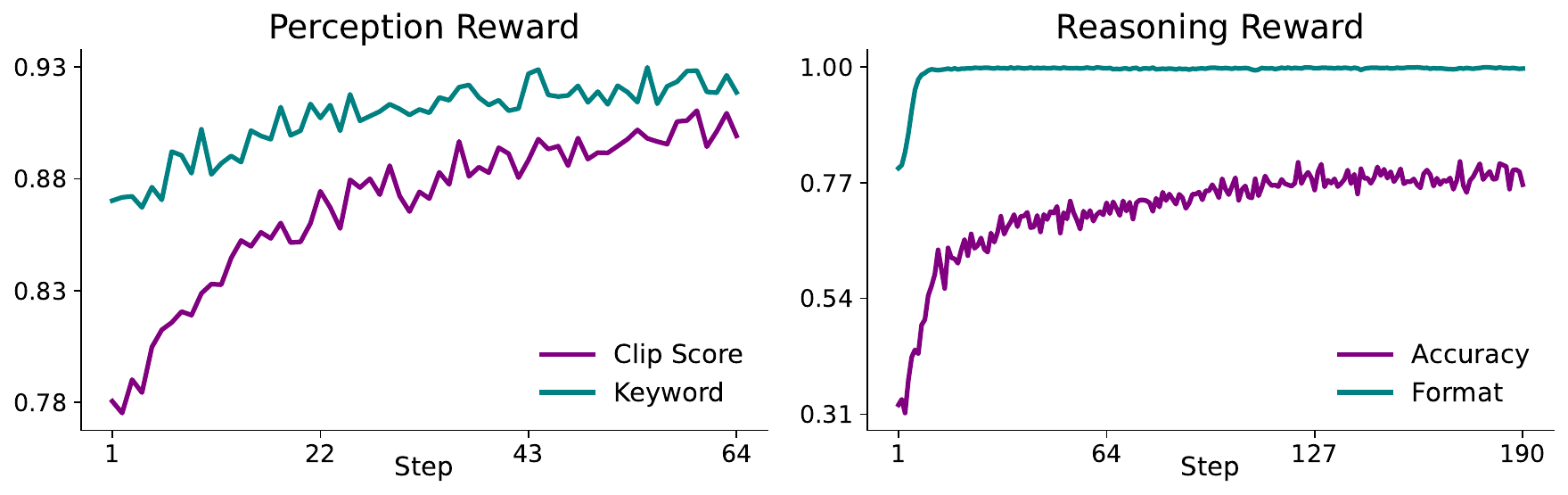}

\caption{Visualization of reward trends. Left: CLIP and keyword rewards during perception-stage training. Right: Accuracy and format rewards during reasoning-stage training.}
\vspace{-0.7em} 
\label{fig4}
\end{figure}

To analyze how model capabilities evolve during training, we visualize reward curves across the two-stage reinforcement learning process using the warm-up-trained Qwen2.5-VL-7B model. As shown in Figure~\ref{fig4}, after the warm-up stage, the model achieves initial CLIP Score and Keyword rewards of around 0.78 and 0.86, indicating basic perception ability. During Perception RL, both rewards steadily increase to approximately 0.91 and 0.93, demonstrating effective enhancement of visual perception. In the Reasoning RL stage, Medium cases are used to ensure stable learning. The reasoning accuracy reward consistently rises, confirming the stage’s effectiveness in improving logical reasoning. Additionally, the Format reward remains close to 1 throughout training, as the supervised fine-tuning imparts the model with a strong understanding of output format.

\section{Conclusion }
We present PeBR-R1, a vision-language model (VLM) trained with a two-stage reinforcement learning strategy, designed to separately enhance both perception and reasoning capabilities. To mitigate the issue of advantage vanishing during training, we design a task difficulty-based sampling strategy that categorizes the dataset into Easy, Medium, and Hard cases, corresponding to Perception RL, Reasoning RL, and supervised fine-tuning, respectively. In the Perception RL stage, we employ both CLIP score and keyword accuracy to improve coarse- and fine-grained visual understanding. In the Reasoning RL stage, we leverage stable and informative reward signals from Medium cases to optimize logical reasoning. PeBR-R1 demonstrates superior performance across seven widely-used benchmarks, validating the effectiveness of our two-stage, task-specific reinforcement learning strategy in bridging the vision-language gap.

\bibliography{aaai2026}

\clearpage
\thispagestyle{plain}

\appendix
\twocolumn[ 
\vspace{1em}
\begin{center}
{\Large\bf Supplementary Material \par}
\end{center}
\vspace{1em}
]
\setcounter{secnumdepth}{2} 
\renewcommand{\thesection}{\Alph{section}}      
\renewcommand{\thesubsection}{\thesection.\arabic{subsection}}
\section{Additional Evaluation on 7B  Models}
Due to space constraints, we report results on seven representative benchmarks. To further validate our model’s reasoning capabilities, we additionally evaluate it on the MathVerse benchmark, comparing against open-source 7B-scale reasoning models, as presented in Table~\ref{table1_appendix}.
\begin{table}[H]
\centering
\begin{tabular}{lcc}
\hline
\textbf{Model}    & \textbf{\begin{tabular}[c]{@{}c@{}}RL \\ Method\end{tabular}} & \textbf{\begin{tabular}[c]{@{}c@{}}MathVerse\\ (testmini)\end{tabular}} \\ \hline
Qwen2.5-VL-7B     & RLHF                                                          & 49.2                                                                    \\
R1-VL-7B          & R-RL                                                          & 40.0                                                                    \\
ThinkLite-VL-7B   & R-RL                                                          & 52.1                                                                    \\
OpenVLThinker-7B  & R-RL                                                          & 47.9                                                                    \\
Vision-R1-7B      & R-RL                                                          & 51.9                                                                    \\
VL-Rethinker-7B   & R-RL                                                          & 54.2                                                                    \\
R1-Share-VL-7B    & R-RL                                                          & 52.8                                                                    \\ \hline
PeBR-R1-7B (Ours) & P+R-RL                                                        & 53.6                                                                    \\ \hline
\end{tabular}
\caption{Evaluation results of 7B-scale models on the MathVerse (testmini) benchmark.}
\label{table1_appendix}
\end{table}

\section{Comparison Between Original and Extended SFT Datasets}
Since the model exhibited noticeable performance degradation in certain aspects after supervised fine-tuning (SFT) on the original Mulberry dataset, we supplemented the SFT data. The following section introduces the sources of the additional data and compares model performance before and after the supplementation.
\subsection{Details of the Additional SFT Samples}
To enhance the diversity and robustness of the SFT data, we added approximately 39K samples from CLEVR-Math, DVQA, TabMWP, VQA2.0, Super-CLEVR, GeoQA+, and a selection of sampled hard cases. Table~\ref{table2_appendix} provides a detailed breakdown of the dataset sizes.
\begin{table}[H]
\centering
\begin{tabular}{c|c}
\hline
Data source & Data size \\ \hline
Clevr\_Math & 9730      \\
DVQA        & 8676      \\
TabMWP      & 4908      \\
VQA2.0      & 3298      \\
Super-CLEVR & 2398      \\
GeoQA+      & 1336      \\
Hard Cases  & 8786      \\ \hline
\end{tabular}
\caption{Data sources and sample sizes of the 39K supplemental SFT dataset.}
\label{table2_appendix}
\end{table}

\subsection{Effect of Dataset Expansion on Model Performance}
After supplementing the training data, the overall performance of the model improved noticeably. Table~\ref{table3_appendix} presents a comparison of Qwen2.5-VL-7B's performance on MathVista using the original Mulberry dataset and the augmented SFT dataset.
\begin{table}[H]
\centering
\begin{tabular}{l|c}
\hline
Method            & Mathvista \\ \hline
Qwen2.5-VL-7B(Baseline)          & 68.2      \\
Origin sft data   & 65.2      \\
With new sft data & 69.8      \\ \hline
\end{tabular}
\caption{Comparison of MathVista performance before and after augmenting the SFT dataset for Qwen2.5-VL-7B.}
\label{table3_appendix}
\end{table}

\section{Visualization of Response Length Dynamics} 
\begin{figure}[H]
\centering
\includegraphics[width=\columnwidth]{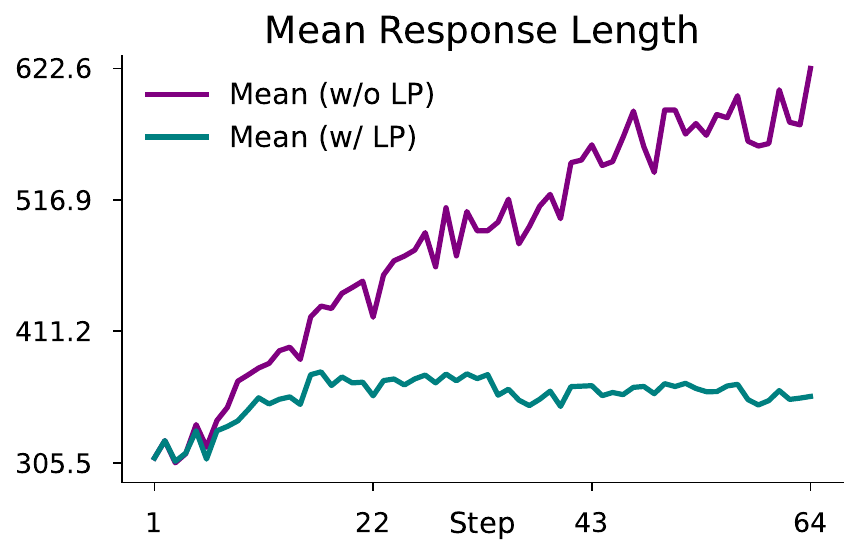}

\caption{Mean response length comparison with and without length penalty during Stage-1 RL.}
\label{fig1_appendix}
\end{figure}

To prevent the model from generating excessively long outputs during the first stage of reinforcement learning (RL), which can degrade performance and destabilize the second-stage training, we introduce a length penalty. Figure~\ref{fig1_appendix} illustrates the average response length at each training step, comparing models with and without the length penalty. As shown, without the penalty, the response length increases approximately linearly with training steps. In contrast, the length penalty effectively suppresses this growth, leading to more stable and controlled outputs.

\section{Comparing GRPO and Supervised Fine-Tuning}
After the warm-up stage, we further conducted a two-stage reinforcement learning process. Since keyword extraction is required in both stages, we first prompted each sample in the RL dataset using Seed1.5-VL to obtain model-generated responses. These responses, combined with the original prompts, can also be used to construct a new SFT dataset for supervised fine-tuning. We take the model after the warm-up stage as the baseline, to compare the effectiveness of SFT and GRPO, we performed both SFT and GRPO training on the same RL dataset to evaluate which method yields greater performance gains. As shown in Table~\ref{table4_appendix}, GRPO leads to more substantial improvements. This is because reinforcement learning better activates and optimizes the model’s perceptual and reasoning abilities—capabilities that conventional SFT alone fails to sufficiently develop.
\begin{table}[H]
\centering
\begin{tabular}{l|c}
\hline
Method   & Mathvista \\ \hline
Warm-up Model(Baseline) & 69.8      \\
GRPO     & 76.0      \\
SFT      & 73.7      \\ \hline
\end{tabular}
\caption{MathVista performance comparison between GRPO and SFT on the same RL dataset.}
\label{table4_appendix}
\end{table}

\section{CLIP Score as a Reward for Perceptual Alignment}
At the coarse-grained level of the perception phase, we introduce a reward mechanism based on the CLIP Score, using FG-CLIP to compute the similarity between the image and its corresponding image description. Built upon the original CLIP, FG-CLIP incorporates phrase-level contrastive learning objectives, significantly enhancing the model’s ability to capture fine-grained details in the image, such as objects, attributes, and relationships. Moreover, the proposed phrase-aware learning framework demonstrates strong generalizability, extending beyond image-text retrieval to other multimodal tasks that demand local semantic understanding, including phrase grounding, visual question answering (VQA), and image captioning.

\begin{figure*}[t]
\centering
\includegraphics[width=\textwidth]{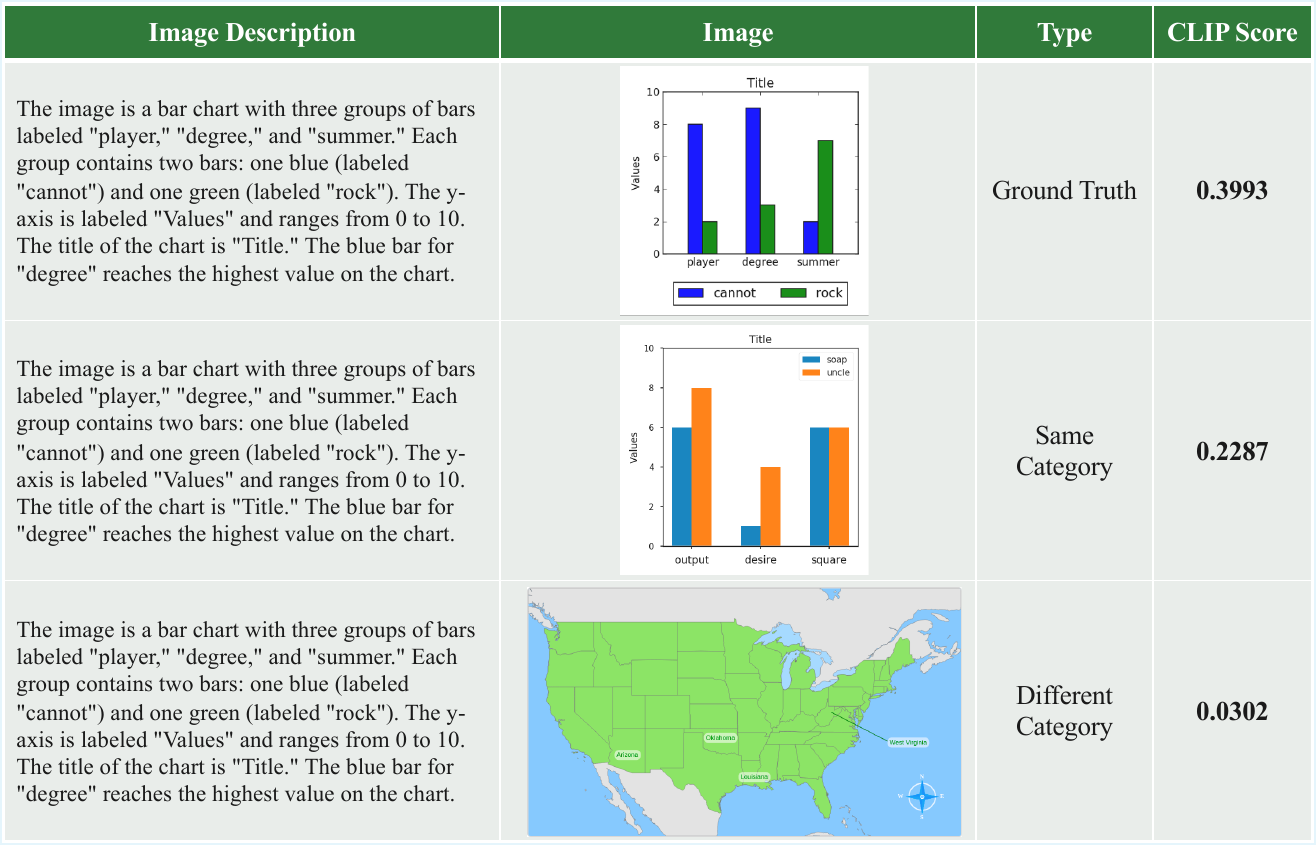}
\caption{FG-CLIP scores across different visual scenarios. The figure illustrates the effectiveness of phrase-aware fine-grained alignment in distinguishing perceptual quality between model-generated descriptions and images.}
\label{fig2_appendix}
\end{figure*}

Therefore, we adopt FG-CLIP as the scoring model to evaluate visual perception quality, aiming to improve the model’s capability in local alignment and fine-grained perception. Figure~\ref{fig2_appendix} presents the CLIP Scores computed by FG-CLIP across different scenarios, illustrating its effectiveness in distinguishing the perceptual quality of model outputs.

The images and image descriptions are sourced from the Mulberry dataset. In the "Type" field, Ground Truth refers to the original image-description pairs from the dataset, while Same Category and Different Category indicate images that belong to the same or different category as the ground truth image, respectively.

\section{Reward Design and Parameter Settings}
The overall reward formulation for our two-stage training is defined in Equations~\ref{eq:reward_stage1} and~\ref{eq:reward_stage2}. In these two stages, the reward weights are set as follows: $\alpha_1 = \beta_1 = 0.4$, $\gamma_1 = \delta_1 = 0.1$, $\alpha_2 = 0.9$, and $\beta_2 = 0.1$.

\begin{equation}
r_1 = \alpha_1 \cdot r_{\text{clip}} + \beta_1 \cdot r_{\text{keyword}} + \gamma_1 \cdot r_{\text{format}} + \delta_1 \cdot r_{\text{length}}
\label{eq:reward_stage1}
\end{equation}

\begin{equation}
r_2 = \alpha_2 \cdot r_{\text{accuracy}} + \beta_2 \cdot r_{\text{format}}
\label{eq:reward_stage2}
\end{equation}

The CLIPScore reward $r_{\text{clip}}$ is defined as follows in Equation~\ref{eq:clip_reward}, where the threshold parameter is set to $\tau_{\text{clip}} = 0.4$:

\begin{equation}
r_{\text{clip}} =
\begin{cases}
1, & \text{if } \text{clipscore} \geq \tau_{\text{clip}} \\
\text{clipscore} / \tau_{\text{clip}}, & \text{otherwise}
\end{cases}
\label{eq:clip_reward}
\end{equation}

To prevent excessively long responses, we introduce a length penalty. The length penalty $r_{\text{length}}$ is computed as follows in Equation~\ref{eq:length_penalty}, where $L_{\text{expected}} = 512$ and $\epsilon$ is a small constant for numerical stability:
\begin{equation}
r_{\text{length}} = \min\left(1.0,\ \frac{L_{\text{expected}}}{L_{\text{actual}} + \epsilon}\right)
\label{eq:length_penalty}
\end{equation}

Since we employ a rule-based reward computation method, the model is required to produce outputs in a predefined structure to ensure consistent evaluation. The expected output format is illustrated as follows:

\vspace{0em}
\begin{quote}
\begin{verbatim}
### Image Description:
### Rationales:
### Let's think step by step.
### Step 1:
### Step 2:
...
### The final answer is: 
\end{verbatim}
\end{quote}
\vspace{0em}

To encourage the model to follow this structure, we introduce a format reward denoted as $r_{\text{format}}$, defined by
\begin{equation}
r_{\text{format}} =
\begin{cases}
1, & \text{if the output follows the predefined format}\\
0, & \text{otherwise}
\end{cases}
\end{equation}
This formulation enforces structural consistency in model responses, facilitating stable and reliable rule-based reward computation.

\section{Representative Cases Before and After Perception RL}
\begin{figure*}[t]
\centering
\includegraphics[width=\textwidth]{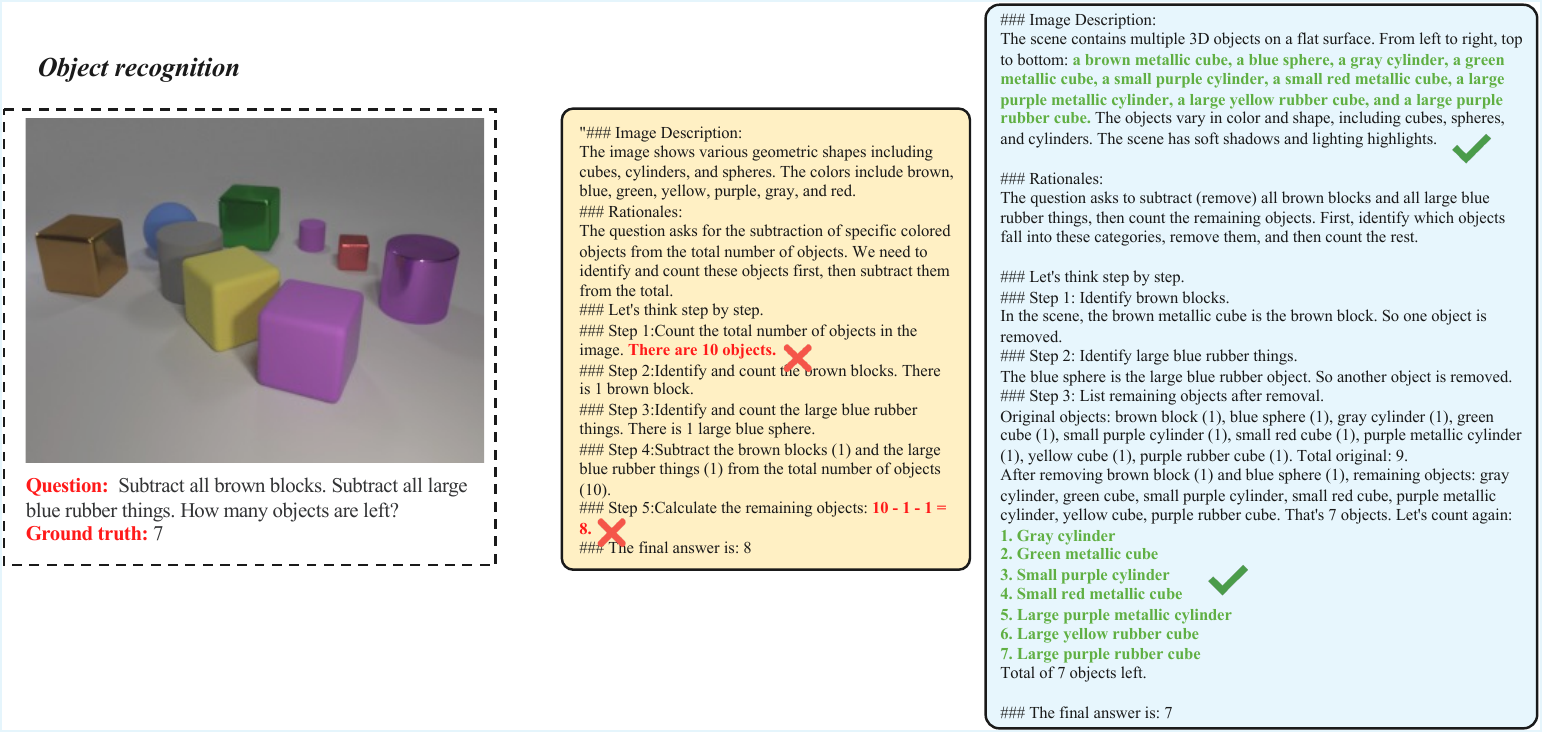}
\caption{Representative Case: Object recognition. The left response is generated by the warm-up Qwen2.5-VL-7B model, while the right response is generated by our PeBR-R1-7B model.}
\label{fig3_appendix}
\end{figure*}

\begin{figure*}[t]
\centering
\includegraphics[width=\textwidth]{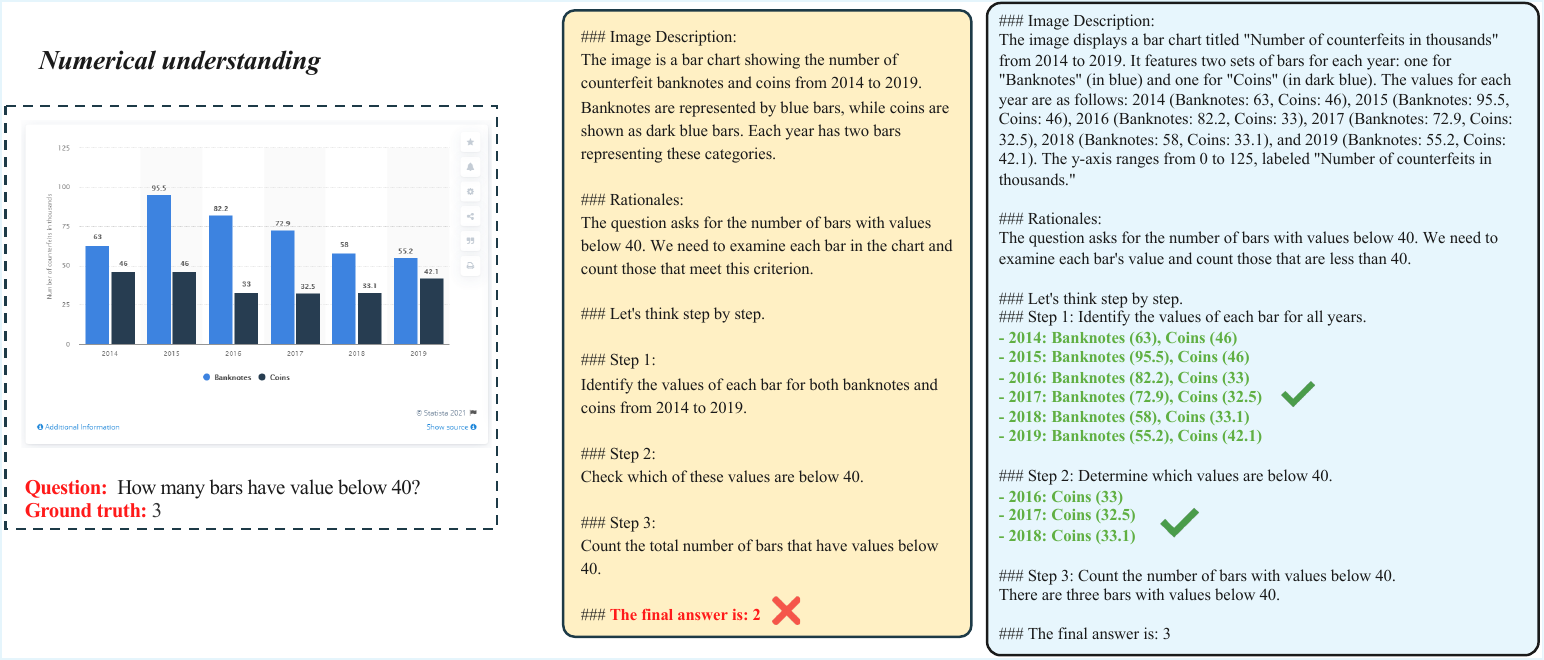}
\caption{Representative Case: Numerical understanding. The left response is generated by the warm-up Qwen2.5-VL-7B model, while the right response is generated by our PeBR-R1-7B model.}
\label{fig4_appendix}
\end{figure*}

\begin{figure*}[t]
\centering
\includegraphics[width=\textwidth]{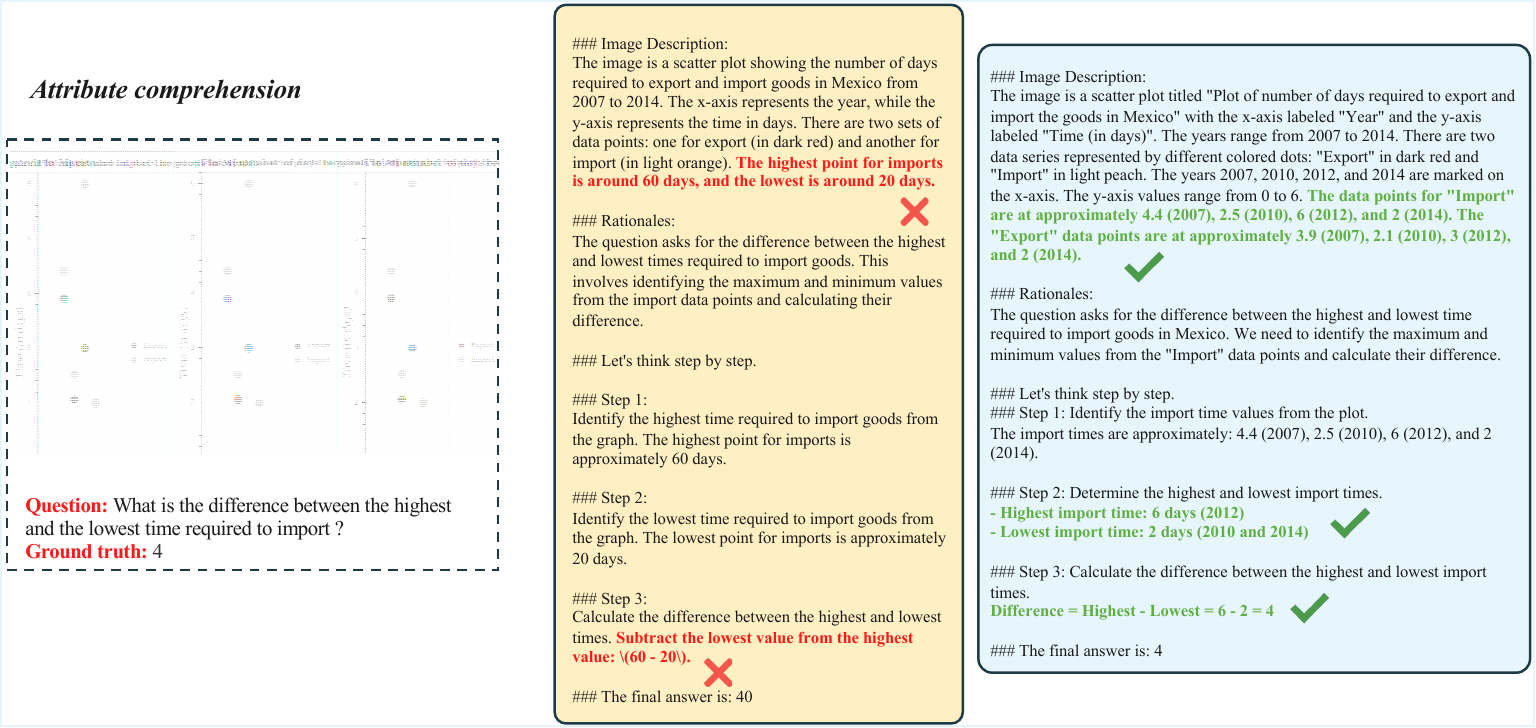}
\caption{Representative Case: Attribute comprehension. The left response is generated by the warm-up Qwen2.5-VL-7B model, while the right response is generated by our PeBR-R1-7B model.}
\label{fig5_appendix}
\end{figure*}

\begin{figure*}[t]
\centering
\includegraphics[width=\textwidth]{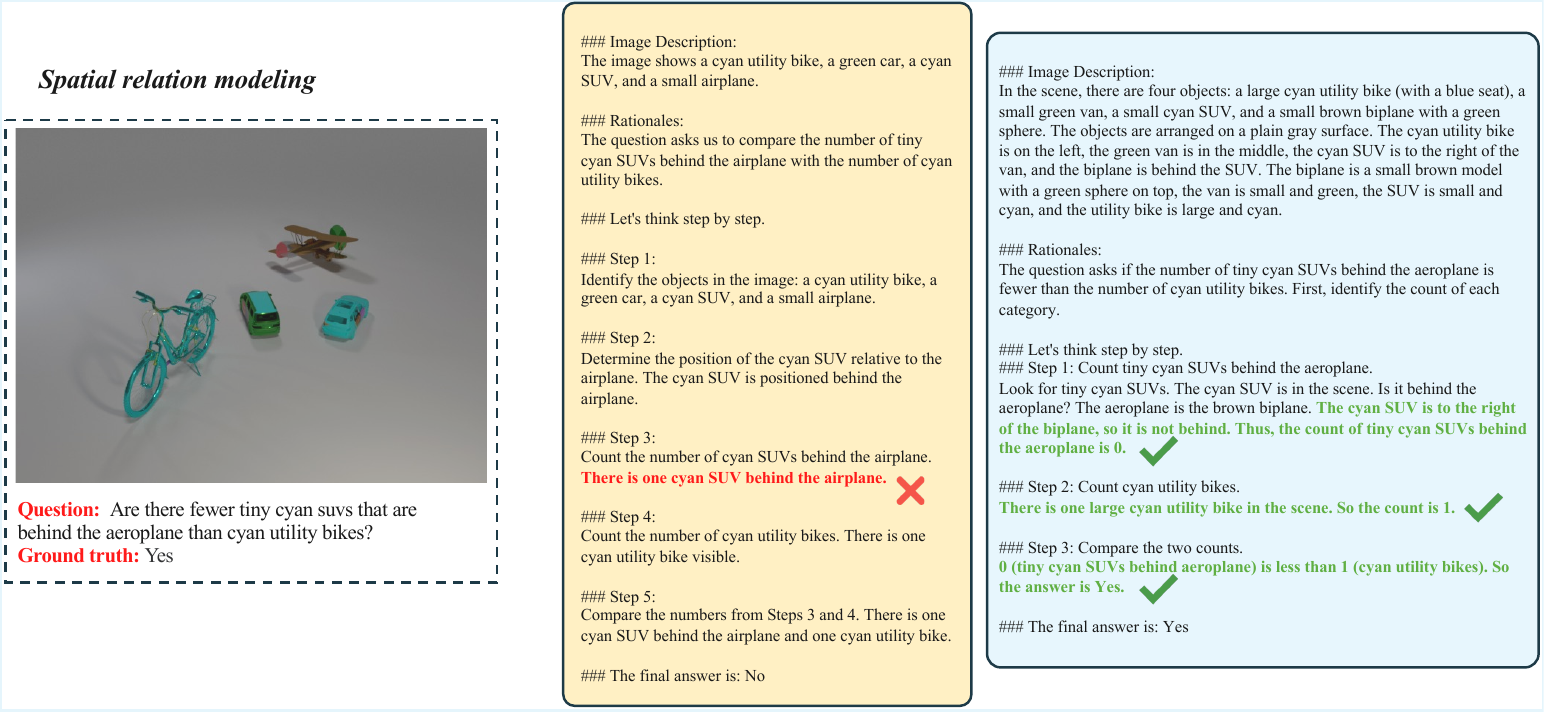}
\caption{Representative Case: Spatial relation modeling. The left response is generated by the warm-up Qwen2.5-VL-7B model, while the right response is generated by our PeBR-R1-7B model.}
\label{fig6_appendix}
\end{figure*}

Our perception reinforcement learning stage is designed to enhance the model’s capabilities in object recognition, numerical understanding, attribute comprehension, and spatial relation modeling. To intuitively illustrate the improvements in perceptual ability before and after reinforcement learning—and how these improvements further facilitate subsequent reasoning—we present several representative examples in Figures~\ref{fig3_appendix}, \ref{fig4_appendix}, \ref{fig5_appendix}, and \ref{fig6_appendix}. These figures compare the outputs of the warm-up Qwen2.5-VL-7B model with those of our enhanced model trained using PeBR-R1. The results clearly demonstrate that PeBR-R1 significantly improves the model’s understanding of visual details and semantic alignment. This enhanced perceptual ability also lays a solid foundation for more accurate and coherent reasoning in later stages.

\end{document}